%% file: fed_mv_mf.tex
\documentclass{article}

\usepackage{arxiv}
\usepackage{hyperref}       
\usepackage{url}            
\usepackage{booktabs}       
\usepackage{amsfonts}       
\usepackage{nicefrac}       
\usepackage{microtype}      

\usepackage{microtype}
\usepackage{graphicx}
\usepackage{subfigure}
\usepackage{booktabs} 
\usepackage{amsmath, amssymb, amsfonts}

\usepackage{hyperref}

\usepackage{siunitx}
\usepackage{mathtools}
\usepackage{cuted}
\usepackage{fancyhdr}
\usepackage{lastpage}
\usepackage[yyyymmdd]{datetime}

\usepackage{setspace}
\usepackage{parskip}
\usepackage{tabu}
\usepackage{multirow}
\usepackage{relsize}

\usepackage{color}
\usepackage{algorithm}
\usepackage{algorithmic}

\usepackage{physics}
\usepackage{authblk}

\newcommand{\p}{{p}}

\newcommand{\Pb}{\textbf{P}}
\newcommand{\Q}{\textbf{Q}}
\newcommand{\Ub}{\textbf{U}}
\newcommand{\V}{\textbf{V}}
\newcommand{\X}{\textbf{X}}
\newcommand{\Y}{\textbf{Y}}

\newcommand{\R}{\textbf{R}}

\author[$\dagger$]{Adrian Flanagan}
\author[ ]{Were Oyomno}
\author[ ]{Alexander Grigorievskiy}
\author[ ]{Kuan Eeik Tan}
\author[$*$]{Suleiman A. Khan}
\author[$*,\dagger$]{Muhammad Ammad-ud-din}
\affil[ ]{Helsinki Research Center, Europe Cloud Service Competence Center}
\affil[ ]{Huawei Technologies Oy (Finland) Co. Ltd}
\affil[$\dagger$]{\textit {Corresponding author:{\{adrian.flanagan,muhammada.din\}}@huawei.com}}
\affil[$*$]{\textit {Equal Contribution}}

\begin{document}

\title{Federated Multi-view Matrix Factorization for Personalized Recommendations}
%
%

\maketitle   




\input{abstract}
\input{introduction}

\input{mvmf_method}
\input{related_works}
\input{datasets}
\input{experiments_results}

\input{conclusion}

\bibliography{ref}
\bibliographystyle{unsrt}

%
%
%

\end{document}

%% file: abstract.tex
\begin{abstract}
	
	We introduce the federated multi-view matrix factorization method that extends the federated learning framework to matrix factorization with multiple data sources. Our method is able to learn the multi-view model without transferring the user's personal data to a central server. As far as we are aware this is the first federated model to provide recommendations using multi-view matrix factorization. The model is rigorously evaluated on three datasets on production settings. Empirical validation confirms that federated multi-view matrix factorization outperforms simpler methods that do not take into account the multi-view structure of the data, in addition, it demonstrates the usefulness of the proposed method for the challenging prediction tasks of cold-start federated recommendations.

\end{abstract}

%% file: introduction.tex
\section{INTRODUCTION}

In many machine learning problems multiple heterogeneous and related data sources or views are available to build a model. The challenge is to effectively integrate the views into a coherent model which performs better than the equivalent single view based model. As an example of a multiple view problem we consider the case of a movie recommender system where historical user-movie watch data allows us to generate user personalized recommendations. By adding other sources of information or data views such as user personal information (e.g., age, gender, and location) and movie features (e.g. for movies genre, actors, director, box office revenue) we would expect to generate better personalized recommendations. While we use the example of a movie recommender the methods described here can be generalized to any type of recommender system. 

With an increasing focus on user privacy and legislation such as the GDPR\footnote{https://gdpr-info.eu/} users may not opt-in to share their personal data and companies may be less willing to record, upload and store users' data to generate multi-view recommendations. The Federated Learning (FL) \cite{mcmahan2017communication} paradigm addresses the issue of users' privacy. In FL model learning is distributed to the end clients (i.e. user's devices), and model updates are generated locally with the users' data and only the model updates are uploaded and aggregated in a central server ensuring the raw user's private data never leaves the client device. 

\begin{figure}[t]
	\centering
	\begin{minipage}{.4\columnwidth}
		\centering
		\includegraphics[width=0.85\textwidth]{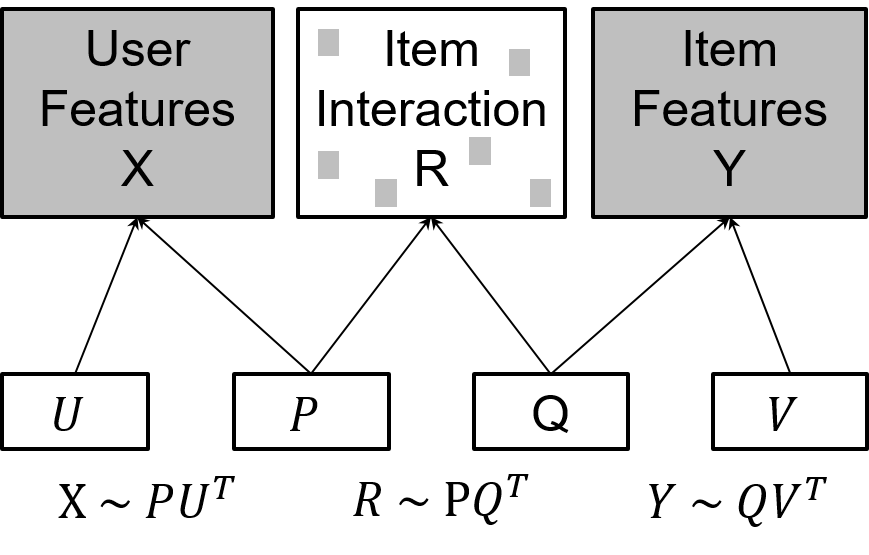}
	\end{minipage}
	\begin{minipage}{.4\columnwidth}
		\centering
		\includegraphics[width=0.85\textwidth]{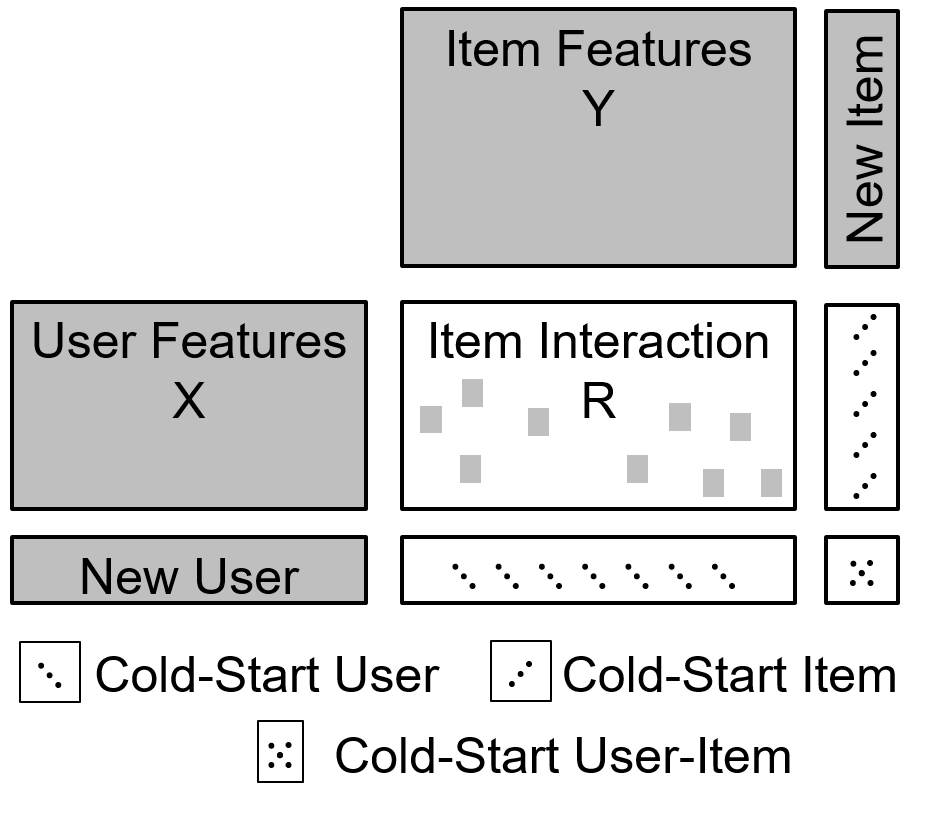}
	\end{minipage}
	\label{fig:overview}
	\caption{Multi-view matrix factorization method with side-information sources (left). Cold-start prediction using FED-MVMF (Right). For prediction of a newly released item, the model predicts the user-item interactions (cold-start item). Similarly, if a new user comes in, the model can make predictions on which items this user is likely to interact, even though there is zero historical interaction data for this user (cold-start user). Finally, when a new user signs up and new items are released, the model can make prediction with zero historical user-item interaction information for the user and item both (cold-start user item).}
\end{figure}

Multiple data views can be combined using Multi-View Matrix Factorization (MVMF) which is an extension of the standard Collaborative Filter (CF)~\cite{balabanovic1997fab,sarwar2001item} for generating recommendations. Figure \ref{fig:overview} conceptualizes the principle of learning from the multiple data views 1) user-item (video), 2) user-features, 3) item (video)-features for predicting personalized movie recommendations. The goal of multi-view matrix factorization is to learn a joint factorization of all the three data matrices. Essentially, the joint factorization decomposes the observed data into sets of low-dimensional latent factors that capture the dependencies between the matrices. The MVMF model therefore uses a combination of the 3 datasets to learn a better model of the user-item interactions. As a result, the model is able to significantly improve recommendations to the user, thereby enhancing user experience. With the blend of matrix factorization, latent variables and multi-view machine learning approaches, it is possible to address several challenges in recommendation systems such as generating a recommendation for a new user (cold-start user), a recommendation for new items (cold-start item) and a recommendation for an entirely new user and new item (out-of-matrix prediction), which is effectively not possible with the simpler CF approaches. %

In what follows we describe a Federated Learning (FL) implementation of Multi-View Matrix Factorization (MVMF). The use of Federated MVMF (FED-MVMF) also allows us to address the issue of cold start in recommender systems in a distributed FL setting. We show that federation of the MVMF is technically feasible and formulate the updates using a stochastic gradient-based approach. We compare our multi-view approach with single-view matrix factorization on the MovieLens, BooksCrossing and an in-house production dataset (anonymized). The findings confirm that our model substantially outperforms the simpler alternatives. In addition, we empirically demonstrate cold-start recommendations with FED-MVMF.
Our original contributions in this work are three fold: (1) we formulate as far as we know the first federated multi-view matrix factorization method with side-information sources, (2) we empirically demonstrate that the method outperforms simpler federated Collaborative Filter methods, (3) we present the first mechanism for cold-start predictions in federated learning mode.

%% file: mvmf_method.tex
\section{Multi-view matrix factorization}
Given multi-view data sources $\R \in \mathcal{R}^{N_u,N_v}$, $\X \in \mathcal{R}^{N_u,D_u}$ and $\Y \in \mathcal{R}^{N_v,D_v}$, for $N_u$ users, $N_v$ items, characterized by $D_u, D_v$ descriptive features, the multi-view matrix factorization method MVMF is defined as a generative model~\cite{fang2011matrix}:

\begin{align}
\R \sim \Pb \Q^T, \\
\X \sim \Pb \Ub^T, \nonumber \\
\Y \sim \Q \V^T \nonumber
\end{align}\label{mvmf_eqn}

where  $r_{ij}$ represents the interaction between user $i$ and item $j$ value $1 \leq i\leq N_u, 1 \leq j \leq N_v$, and $x_{i,d_u}$ denote the value of user $i$'s personal data feature $d_u$ where $1 \leq d_u \leq D_u$ and $y_{j,d_v}$ denote the value of item $j$ at feature $d_v$, for $1 \leq d_v \leq D_v$. The interactions $r_{ij}$ are generally derived from explicit user feedback such as ratings $r_{ij} \in (1, \ldots, 5)$ given by a user $i$ to an item $j$ \cite{zhou2008large}, or implicit feedback $r_{ij}\geq1$ when the user $i$ interacted with the item $j$ and is unspecified otherwise \cite{Hu2008}. In this work, we consider the case of implicit feedback for simplicity, however, the proposed method is applicable to explicit feedback scenario without loss of generality.

The model learns the latent matrices which are represented as $\Pb \in \mathcal{R}^{N_u,K}$ the user-factor matrix, $\Q \in \mathcal{R}^{N_v,K}$ the item-factor matrix, $\Ub \in \mathcal{R}^{D_u,K}$ user-feature factor matrix and $\V \in \mathcal{R}^{D_v,K}$ the item-feature factor matrix, where $K$ is the number of latent factors.

The shared user-factor matrix $\Pb$ captures the statistical dependencies between the item interaction and user personal data sets as shown in Figure~ \ref{fig:overview}. Likewise, $\Q$ item-factor matrix captures the common patterns between item interactions and item-features data source. The joint factorization, therefore, learns the shared dependencies between the item interactions and side-information sources.
The latent factors $\Ub$ and $\V$ are specific to the user personal and item features respectively, capturing the source-specific variation.

The inference is then performed by optimizing the cost function on the joint factorization of all the data sources as:  
\begin{multline}
J = \sum_{i}\sum_{j} c_{i,j}(r_{i,j} - {p_{i}q_{j}^T})^2 + \lambda_{1} \big( \sum_{i} \sum_{d_{u}} \big( x_{i, d_u} - {p_{i} u_{d_u}^T} \big)^2 + \sum_{v} \sum_{d_y} \big( y_{j,d_y} - {q_{j}v_{d_y}^T} \big) ^2 \big) \\+ \lambda_{2} \big( \sum_{j} ||p_{i}||^2 + \sum_{i}||q_{j}||^2 + \sum_{d_u} ||u_{d_u}||^2 + \sum_{d_y} ||v_{d_v}||^2 \big) 
\end{multline}\label{multview_model_eq}

where $c_{ij} = 1 + \alpha r_{ij}$, for $\alpha >0$ is a confidence parameter to account for implicit feedback uncertainty~\cite{Hu2008} and $\lambda_{(2)}$ are the L2-regularization terms. Specifically, $\lambda_{1}$ can be tuned to adjust the strength of information sharing with the side-data matrices. For example, initializing $\lambda_{1} = 0$ restricts the model to not learn any shared factors, while $\lambda_{1} = 1$ pushes the model to learn factors shared between item interactions and side-data sources. The $0 =< \lambda_{1} =< 1$ value can be chosen informatively based on the prior knowledge about the data generation process or through hyper-parameter optimization. The latent factors $\Pb$, $\Q$, $\Ub$ and $\V$ are inferred using Alternating Least Square \cite{fang2011matrix}.

Several related formulations for joint factorization has been proposed earlier with linear, kernelized and Bayesian variants of matrix factorization with data sources on both sides~\cite{ammad2014integrative,cortes2018cold,gonen2013kernelized,singh2008relational}. 

\section{Federated multi-view matrix factorization}\label{sec:fed_mvmf}
\begin{figure}[ht]
	\centering
	\includegraphics[width=1\columnwidth]{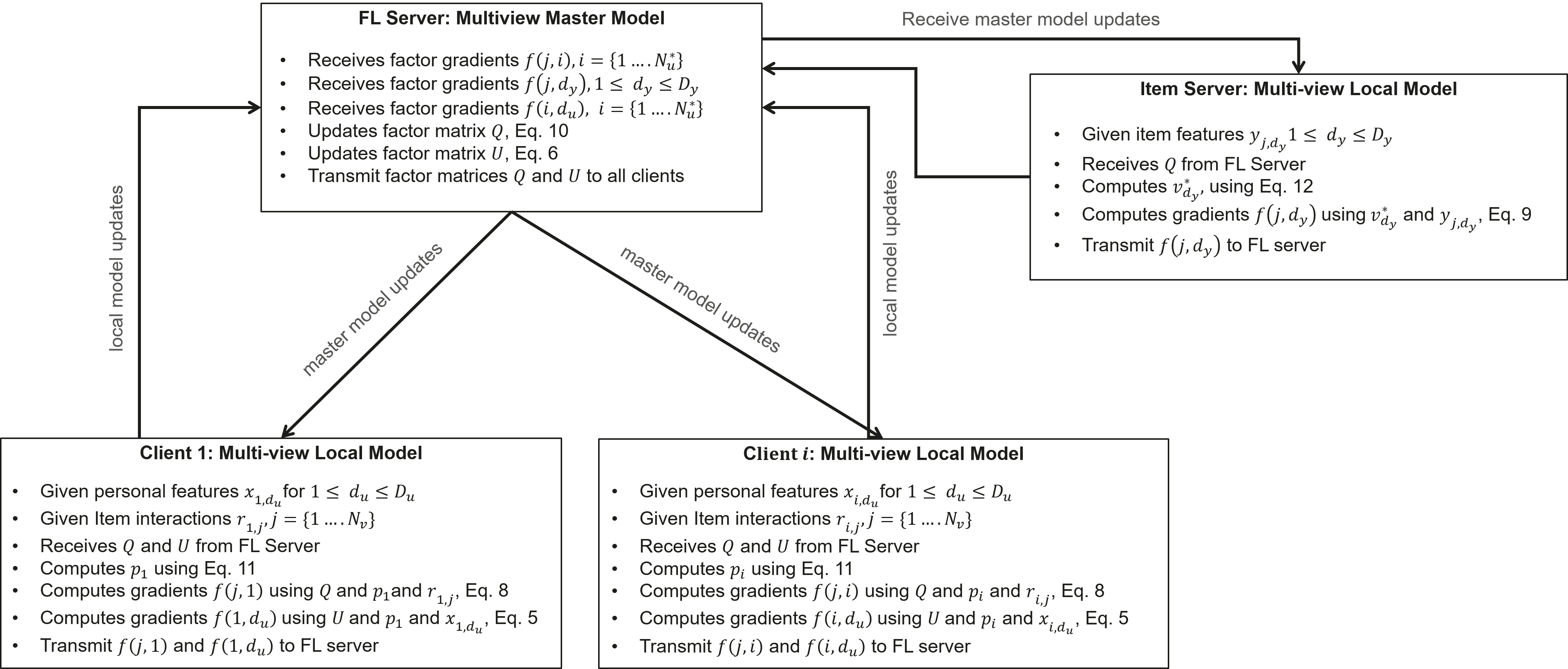}
	\label{fig:fmvmf}
	\caption{Federated multi-view matrix factorization method with side-information data sources. The master model $\Q ,\ \Ub$ is updated on the server and then distributed to users. Each user-specific local model $\p_i$ (user-factor) remains on the user, and is updated on the user using the local user data and $\Q ,\ \Ub$ from the server. The updates through the gradients of $\Q ,\ \Ub$ are computed on each user and transmitted to the server where they are aggregated to update the master model $\Q ,\ \Ub$. Meanwhile, the master model $\Q$ is also transmitted to the item server, item-feature factor matrix $\V$ is updated using the item features. The updates comprising the gradients of $\Q$ are further computed and transmitted to the FL server.}
\end{figure}

We present the first Federated treatment of Multi-view Matrix Factorization (FED-MVMF) which combines side-information sources simultaneously from both sides. The multi-view data sources are distributed and not stored on central servers. The user-item interaction data and user personal data are available on user's devices only, while, the item features are stored on the item server as shown in Figure~\ref{fig:fmvmf}. The proposed framework is presented for the particular case of personalized federated recommendations, though is applicable to other domains as well. 

FED-MVMF performs a federated factorization of the data matrices $\R$, $\X$, $\Y$ jointly as defined in Eq.~\ref{mvmf_eqn} to learn the latent factors $\Pb$, $\Q$, $\Ub$ and $\V$. The federated factorization is formulated using stochastic gradient decent inference. We observe that federated inference of $\Ub$ and $\Q$ is fundamentally challenging as their updates depend on all the users while federated constraints prohibit a direct integration of user-level data. Next, we discuss our federated solution.

\paragraph{\textbf{Federated $\Ub$:}} Federated inference of the user-feature factors $\Ub$ is a key challenge. 
The update requires the user-factor vectors $p_i$ $\forall i\in \{1, \ldots, N_u\}$ from all the users, as
\begin{equation}\label{update_u}
u_{d_u}^{*} = (x_{d_u} P){(P^TP + \frac{\lambda_{2}}{\lambda_{1}}I)}^{-1}. 
\end{equation}
Therefore, $u_{d_u}$ cannot be inferred on individual users and must be carried out on the FL server. However, owing to the privacy constraint, each user also preserves the corresponding $p_i$ locally on the device and can not transmit it to the FL server, further complicating the update of $u_{d_u}^*$.

We formulate a stochastic gradient descent approach to allow for the update of the $u_{d_u}$ vectors on the server, while preserving the user's privacy. Formally, $u_{d_u}$ is updated on the FL server as

\begin{equation}
u_{d_u} \ = \ u_{d_u} \ - \ \gamma \frac{\partial J}{\partial u_{d_u}}, 
\label{fed_derv_u_1}
\end{equation}

for some gain parameter $\gamma$ to be determined. However, computation of  $\partial J/\partial u_{d_u}$ requires a summation over all users. Therefore, we define $f(i, d_u)$ as

\begin{equation}
f(i,d_u) \ = \ \big[(x_{i,{d_u}} - p_i u_{d_u}^T)\big] p_i, 
\label{fed_derv_u_2}
\end{equation}

where $f(i,d_u)$ is calculated on each user $i$ independently of all the other users. All the users then send back the gradient values $f(i,d_u), \ i=\{1,\ldots, N_u\}$ to the FL server. Finally, $\partial J/\partial u_{d_u}$ be formulated as aggregate of the user gradients as
\begin{equation}
\frac{\partial J}{\partial u_{d_u}} = -2 \sum_{i} f(i, d_u) + 2 \lambda_{2} u_{d_u},
\label{fed_update_u}
\end{equation}
enabling federated update of $u_{d_u}$.

\paragraph{\textbf{Federated $\Q$:}}
Analogous to $\Ub$, the federated inference of $\Q$ is also non-trivial and more complex as $\Q$ is a shared factor between user-item interactions and item-features. The inference depends on both latent factors $\Pb$ and $\V$. 
Practically, the update of item factor vectors $q_j \ \forall \ j \in \{1, \ldots, N_v\}$ requires the user factor vectors $p_i$ $\forall i \in \{1, \ldots, N_u\}$ from all the users and $v_{d_y} \ \forall \ v \in \{1, \ldots, N_v\}$ from all the items, 
\begin{equation}\label{update_q}
q_{j}^{*} = (r(j)\hat{C}^{(j)}P + \lambda_{1}y_{j}V)(P^TC^{(j)}P + \lambda_{1}V^TV + \lambda_{2}I)^{-1}. \nonumber
\end{equation}
Therefore, the updates of $q_{j}$ can not be done on the user's device and must be performed on the FL server.
But, due to the privacy constraints, each user preserves the $p_i$ locally on the device and can not send it to the FL server, further complicating the inference of $q_{j}^*$. We present a stochastic gradient descent approach to allow for the update of the $q_{j}$ vectors on the FL server, while preserving the user's personal data.

Formally, $q_{j}$ is updated on the FL server as
\begin{equation}
q_{j} \ = \ q_{j} \ - \ \gamma \frac{\partial J}{\partial q_{j}}, 
\label{fed_derv_q_1}
\end{equation}
for gain parameter $\gamma$ to be determined. However, computing $\partial J/\partial q_{j}$ involves a summation over all users $i$ and item features $d_y$. Therefore, we define $f(j, i)$ and $f(j, d_y)$ as
\begin{equation}
f({j}, i) \ = \ \big[c_{ji} (r_{j,i} - {p_i} q_{j}^T)\big] p_i, 
\label{fed_derv_q_2}
\end{equation}
\begin{equation}
f({j}, d_y) \ = \ \big[(y_{j,d_y} - {v_{d_y}} q_{j}^T)\big] v_{d_y}, 
\label{fed_derv_q_3}
\end{equation}

where $f(j, i)$ is calculated on each user $i$ independently of all the other users. All the users then report back the gradient values $f(j, i), \ i=\{1,\ldots, N_u\}$ to the FL server. And $f({j}, d_j)$ is calculated on the item server and the gradient values $f(j, {d_y}), \ d_y=\{1,\ldots, D_y\}$ are transmitted to the FL server. Finally, the derivative can then be computed using an aggregate of the user and item-features gradients as
\begin{equation}
\frac{\partial J}{\partial q_j} \ = \ -2 \sum_{i} f(j, i) \ - 2 \lambda_1 \sum_{d_y} f(j, d_y) \ + \lambda_{2} q_{j},
\label{fed_update_q}
\end{equation} 
making it possible to perform federated update of $q_{j}$.

\paragraph{\textbf{Localized $\Pb$:}}
The inference of user factors $\Pb$ depends on the item-factors $\Q$, user-features factors $\Ub$ and user's own data $r_{ij}$, locally available at each user. The factor models $\Q$ and $\Ub$ are received from the FL server and are used to compute the corresponding $\p_i^*$ at each user's device as
\begin{equation}\label{update_p}
\p_i^* \ = (r(i)\hat{C}^{(i)}Q + \lambda_{1}x_{i}U)(Q^TC^{(i)}Q + \lambda_{1}U^TU + \lambda_{2}I)^{-1} 
\end{equation}
where $\p_i = \p_{i^*}$ is the optimal solution obtained $\partial J(p_{u^*})/\partial p_u = 0$, from Eq.~\ref{multview_model_eq}. Notably, the updates can be carried out independently for each user $i$ without reference to any other user's personal data.

\paragraph{\textbf{Localized $\V$:}}
The FL server transmits the latest item factors $\Q$ to the item server in each iteration. The item's features $y_{d_y}$ is used to compute the $v_{d_y}^*$ locally for each item. The updates can be carried out independently for each item $j$ without reference to any other private data as
\begin{equation}\label{update_v}
v_{d_y}^{*} = (y_{d_y} Q){(Q^TQ + \frac{\lambda_{2}}{\lambda_{1}}I)}^{-1}. 
\end{equation}
where $v_{d_y} = v_{d_y}^{*}$ is the optimal solution obtained $\partial J(v_{d_y^{*}})/\partial v_{d_y} = 0$, from Eq.~\ref{multview_model_eq}

\begin{algorithm}[ht]
	\small
	\caption{FED-MVMF: Federated Multi-View Matrix Factorization}
	\label{alg:fed_mvmf}
	\begin{algorithmic}[1]
		\STATE \textbf{FL Server}
		\STATE Number of items $N_v$, Number of user features $D_u$, Number of factors $K$
		\STATE Initialize master model factor matrices $\Q$,$\Ub$ and update threshold $\Theta$
		\WHILE{True}
		\STATE Transmit $\Q$ and $\Ub$ to users $i \in [1,N_u]$
		\STATE Transmit $\Q$ $\rightarrow$ \textbf{Item Server}
		\STATE Receive factor $\Q$ gradients $f(j, i)\ \forall \ j \in [1,N_v], \ \forall \ i \in [1,N_u]$
		\STATE Receive factor $\Q$ gradients $f(j, d_y)$ for $d_y \in \ 1 \leq d_y\leq D_y,\ \forall \ j \in [1,N_v]$
		\STATE Receive factor $\Ub$ gradients $f(d_u, i)$ for $d_u \in \ 1 \leq d_u\leq D_u,\ \forall \ i \in [1,N_u]$
		\IF{$\mbox{NumberGradientUpdates}>=\mbox{$\Theta$}$}
		\STATE Update $\Ub$ using \textbf{Eq.\ref{fed_update_u}}
		\STATE Update $\Q$ using \textbf{Eq.\ref{fed_update_q}}
		\ENDIF
		
		\ENDWHILE
		\STATE
		\STATE \textbf{FL User}
		\WHILE{True}
		\STATE Receive master model factor matrices $\Q,\ \Ub$
		\STATE Compute local model factor $p_i$ using Eq.~\ref{update_p}
		\STATE Generate recommendations: $r_{i,j} = \p_i \times \Q^T \ \forall \ j \in [1,N_v]$ 
		\STATE Compute factor $\Q$ gradients $f(j, i)$ using \textbf{Eq.~\ref{fed_derv_q_2}}
		\STATE Compute factor $\Ub$ gradients $f(d_u, i)$ using \textbf{Eq.~\ref{fed_derv_u_2}}
		\STATE Transmit $f(j, i)$ and $f(d_u, i)$ $\rightarrow$ \textbf{FL Server}
		\ENDWHILE
		\STATE
		\STATE \textbf{Item Server}
		\WHILE{True}
		\STATE Receive master model factor matrix $\Q$
		\STATE Compute local model factor $v_{d_y}^{*}$ using \textbf{Eq.~\ref{update_v}}
		\STATE Compute factor $\Q$ gradients $f(j, d_y)$ using \textbf{Eq.~\ref{fed_derv_q_3}}
		\STATE Transmit $f(j, d_y)$ $\rightarrow$ \textbf{FL Server}
		\ENDWHILE
	\end{algorithmic}
\end{algorithm}

A constant gain factor $\gamma$ is used for the update of $Q$ and $\Ub$ and it is seen that the value needs to be chosen with some care to ensure convergence.  Gradient descent is the simplest form of optimisation and there are many variations on it which can lead to faster convergence and greater stability some of which are summarised in \cite{DBLP:journals/corr/Ruder16}. The Adaptive Moment Estimation (Adam) method \cite{kingma2015adam} has also been used in the context of FCF~\cite{ammad2019federated}. We resort to the same approach for inference of FED-MVMF model. In Adam the gradient descent is split into 2 separate parts which record an exponentially decaying of past squared gradients $v_t$ and an exponentially decaying average of past gradients $m_t$,
\begin{equation}
m \ = \ \beta_1 m \ + \ (1 - \beta_1) \frac{\partial J}{\partial q_j} 
\end{equation}
and 
\begin{equation}
v \ = \ \beta_2 v \ + \ (1 - \beta_2) \Bigg(\frac{\partial J}{\partial q_{j}}\Bigg)^2 
\end{equation} 
with $0 < \beta_1, \beta_2 < 1$. The $m, v$ are typically initialized to $0$ values and hence biased towards $0$. To counteract these biases bias corrected versions of $m, v$ are given by,
\begin{equation}
\hat{m} = \frac{m}{1 - \beta_1},
\end{equation}
and
\begin{equation}
\hat{v} = \frac{v}{1 - \beta_2}.
\end{equation}
The updates are then given by,
\begin{equation}
q_j \ = \ q_j - \frac{\gamma}{\sqrt{\hat{v}} + \epsilon}\hat{m}
\end{equation}
$0 < \gamma < 1.0$ is a constant learning rate and $0 < epsilon \ll 1$ e.g. $10^{-8}$ is to avoid a divide by $0$ scenario. We used Bayesian Optimization~\cite{snoek2012practical} approach to chose the values of $\beta_1, \beta_2, \gamma$ and $\epsilon$. 

Likewise, we adopt the same treatment for the inference of user-feature factors $\Ub$

\textbf{Iterative federated updates:}
We outline the steps of the proposed FED-MVMF model in Algorithm~\ref{alg:fed_mvmf}. The FL iterations are performed till the model has converged, where in each iteration master model is updated when the number of collected federated updates from users and item server reached a certain threshold $\Theta$. In the standard mode, the computational complexity of the algorithm is $\mathcal{O}(IK{^2}N_{v}N_{u})$ where $I$ is the number of iterations. However in federated learning set-up, several other parameters can influence the computational complexity of the algorithm such as number of users participating in the update, how frequent the updates are sent by the users, what are specifications of user's devices (laptop or mobile) and importantly the communication over Internet and network latency~\cite{li2019federated}. In Section~\ref{sec:exp_results}, we give payload estimates on the algorithm complexity in terms model sizes and time when tested in production settings.
\section{Federated cold-start recommendations}
The multi-view matrix factorization allows for the inclusion of side-information sources for both users and items simultaneously, making it possible to solve the difficult task of predicting recommendation to new users (cold-start users) or new items (cold-start items) and/or predicting recommendations to an entirely new user on a previously unseen item. Here, a common assumption is that for a new user or a new item, there exist no historical interaction data $r_{*}$, though user's personal features $x_{d_u}^{\text{*}}$ or item's features $y_{d_y}^{\text{*}}$ are available. Figure~\ref{fig:overview} shows the schematic for numerous cold-start recommendation scenarios using standard multi-view matrix factorization. However, in contrast to the cold-start recommendation solution offered by standard approaches~\cite{ammad2014integrative,cortes2018cold}, the FL requires customized solution owing to the privacy constraints and decentralized nature of the multi-view data. We next present the solution of federated cold-start recommendations problem utilizing the proposed FED-MVMF model. 
\paragraph{\textbf{Cold-start user recommendation:}} When a new user joins a FL recommendation system with no previous item interaction data, a new local factor model $\p_{*}$ is created using $\Ub$ and user personal features $x_{d_u}^{\text{*}}$. The cold-start user recommendation is then generated as outlined in Algorithm~\ref{alg:fed_cs_user}.
\begin{algorithm}[ht]
	\small
	\caption{Federated cold-start user recommendation}
	\label{alg:fed_cs_user}
	\begin{algorithmic}[1]
		\STATE \textbf{FL User}
		\STATE Receive master model factor matrices $\Q$,$\Ub \leftarrow$ \textbf{FL Server}
		\STATE Get personal features $x_{d_u}^{\text{*}}$ of new user 
		\STATE Compute new local factor matrix $\p_{*} = x_{d_u}^{\text{*}} \Ub^T$
		\STATE Compute recommendations $r_{*} = \p_{*} \Q^T$ 
	\end{algorithmic}
\end{algorithm}
\paragraph{\textbf{Cold-start item recommendation:}} New items are frequently added to the collection and it is greatly important for the service provider to recommend the new item to a potentially interested user from day zero. The FED-MVMF solves the cold-start item recommendation challenge by creating a new item factor matrix $q_{*} = y_{d_y}^{\text{*}} \V$ at the item server, given the item features $y_{d_y}^{\text{*}}$ and $\V$. The master model item-factor matrix is updated as $\Q_* = [\Q^{(N_v \times K)}\ |\ q^{(1 \times K)}_{*}]$ and transmitted to the FL server. The users receive the updated $\Q_*$ and compute recommendations: $r_{i,*} = \p_{i} \Q_{*}^T$ including the new item as outlined in Algorithm~\ref{alg:fed_cs_item}.
\begin{algorithm}[ht]
	\small
	\caption{Federated cold-start item recommendation}
	\label{alg:fed_cs_item}
	\begin{algorithmic}[1]
		\STATE \textbf{Item Server}
		\STATE Receive master model factor matrix $\Q  \leftarrow$ \textbf{FL Server}
		\STATE Get local item-feature factor matrix $\V$ 
		\STATE Get item features $y_{d_y}^{\text{*}}$ of new item 
		\STATE Compute new item factor matrix $q_{*} = y_{d_y}^{\text{*}} \V$
		\STATE Update the item-factor model matrix $\Q$ with $q_{*} \rightarrow \Q_*$
		\STATE Transmit $\Q_* \rightarrow$ \textbf{FL Server}
		\STATE \textbf{FL Server}
		\STATE Receive updated master model factor matrix $\Q_*$
		\STATE Transmit $\Q_*$ master model to existing users
		\STATE \textbf{FL User}
		\STATE Compute recommendations $r_{i,*} = \p_{i} \Q_{*}^T$
	\end{algorithmic}
\end{algorithm}
\paragraph{\textbf{Cold-start user-item recommendation:}}
The prediction of cold-start user-item recommendation is deemed as out-of-matrix prediction task and is perhaps the most challenging in practice. However, the solution is made possible by FED-MVMF with inclusion of factor matrices originating from side-information sources. Technically, FED-MVMF solves the prediction task by combining solutions of \textit{federated cold-start user recommendation} and \textit{federated cold-start item recommendation}. The user creates a new local factor model $\p_{*}$ using $\Ub$ and user personal features $x_{d_u}^{\text{*}}$ and receives the updated master model item-factor matrix $\Q_*$ to compute recommendations: $r_{*} = \p_{*} \Q_{*}^T$ for the new item as outlined in Algorithm~\ref{alg:fed_cs_user_item}.
\begin{algorithm}
	\small
	\caption{Federated cold-start user-item recommendation}
	\label{alg:fed_cs_user_item}
	\begin{algorithmic}[1]
		\STATE \textbf{Item Server}
		\STATE Receive master model factor matrix $\Q  \leftarrow$ \textbf{FL Server}
		\STATE Get local item-feature model matrix $\V$ 
		\STATE Get item features $y_{d_y}^{\text{*}}$ of new item 
		\STATE Compute new item factor matrix $q_{*} = y_{d_y}^{\text{*}} \V$
		\STATE Update the item-factor model matrix $\Q$ with $q_{*} \rightarrow \Q_*$
		\STATE Transmit $\Q_* \rightarrow$ \textbf{FL Server}
		\STATE \textbf{FL Server}
		\STATE Receive updated master model factor matrix $\Q_*$
		\STATE Transmit $\Q_*, \Ub$ master model to existing users
		\STATE \textbf{FL User}
		\STATE Receive master model factor matrices $\Q_*, \ \Ub$
		\STATE Get personal features $x_{d_u}^{\text{*}}$ of new user 
		\STATE Compute new local factor matrix $\p_{*} = x_{d_u}^{\text{*}} \Ub^T$
		\STATE Compute recommendations $r_{*} = \p_{*} \Q_{*}^T$ 
	\end{algorithmic}
\end{algorithm}

%% file: related_works.tex
\section{Related work}\label{sec:related_works}
The federated multi-view matrix factorization problem and our solution for it are related to several matrix factorization as well as federated learning methodologies. We next discuss the existing methods that solve special cases of the problem, and relate them to our work.
\subsection{Multi-view learning}
For the non-federated case, our model can be seen as a multi-view matrix factorization with side-information sources~\cite{fang2011matrix}.

\textbf{One-way factorization:} 
Several methods perform integrated analysis of multiple matrices $\mathbf{X}^{(1)} \in \mathcal{R}^{N \times D_1}$, ..., $\mathbf{X}^{(M)} \in \mathcal{R}^{N \times D_M}$, where $N$ is the number of paired samples in $M$ matrices with $D_m$ dimensions, such that the matrices are paired in one mode only. Classical approaches like Canonical Correlation Analysis~\cite{hotelling1936relations} perform joint factorization of two matrices. More recent advancements, like Group Factor Analysis~\cite{klami2015group} can integrate several matrices. However, unlike ours, none of these methods perform factorization of matrices paired on both sides.

\textbf{Two-way factorization:}
Similar to our approach a few methods perform two-way factorization of matrices coupled in both modes~\cite{cortes2018cold,fang2011matrix}. Moreover, \cite{gonen2013kernelized} introduced a non-linear Kernelized Bayesian Matrix Factorization coupled with multiple side-information sources in $\X$ and $\Y$. Recently, \cite{bunte2016sparse} extended the group factor analysis approach for bi-clustering using two-way factorization that integrates data sources from both sides.
Extending the similar line of research, \cite{strahl2020} scales matrix factorization with two-way side information sources for efficient inference when the number of covariates is large. Two way factorization or collective matrix factorization methods are especially suitable and widely used for personalized recommendation applications \cite{ammad2014integrative,cortes2018cold,singh2008relational}. 

However, none of these methods presents a federated learning solution, and our method is the first to provide a federated multi-view matrix factorization integrating side-information sources from both sides.
\subsection{Federated learning}
Within federated learning, our method is a general multi-view matrix factorization, where several existing methods can be seen as special cases of our model.

\textbf{One matrix:}
For a single matrix, our model can be seen as a collaborative filter \cite{ammad2019federated,chai2019secure,dolui2019poster}. This case is also close to distributed matrix factorization of \cite{gemulla2011large,yu2014parallel,zhou2008large}. However, none of these approaches is able to integrate multiple data sources.

\textbf{Two matrices:}
For two data sets, with partially paired samples, vertical federated learning approaches \cite{hardy2017private,liu2018secure} take advantage of the common samples, while horizontal federated learning approaches \cite{mcmahan2017communication,DBLP:conf/nips/SmithCST17} leverage the overlap of feature columns to improve the predictive performances. For a comprehensive review of these approaches see \cite{li2019federated,yang2019federated}.
Furthermore, recently~\cite{huangiterative} performed federated factorization of multiple matrices, however, their method is able to factorize matrices paired in a single mode only.

Other federated learning approaches based on neural networks \cite{DBLP:journals/corr/KonecnyMRR16,mcmahan2017communication} do not address the problem of personalized recommendation using multiple side-data sources. In addition, \cite{mcmahan2017communication} aggregates model weights at the server whereas we employ a gradient based aggregation suitable for matrix factorization. Other approaches like meta-learning~\cite{chen2018federated} have been proposed in the context of recommendation systems. Recently, \cite{jalalirad2019simple} adapt the meta-learning approach and parallel implementation of federated learning.  More recently, \cite{bonawitz2019towards} discuss optimizing federated learning at scale, however, none of these methods address the multi-view matrix factorization problem.

To the best of our knowledge, our method is the first federated multi-view matrix factorization that integrates data from both sides to provide personalized federated recommendations.

%% file: datasets.tex
\section{DATA}
In this study, we used three datasets: two public and a private in-house anonymized production dataset. These datasets are \textit{MovieLens-1M}~\cite{harper2016movielens}, \textit{BookCrossings}~\cite{ziegler_2005} and \textit{in-house} (anonymized). These datasets are characterized by varying degree of sparse user-item interactions, in addition to having descriptive features for both users and items. Interactions of in-house dataset are implicit while interactions of public datasets are explicit i. e. they include an exact rating a user specified for an item. For generalize treatment, we convert the public datasets into implicit datasets as well. The pre-processing details of each dataset are provided in the Supplementary Material. Final characteristics of a dataset after pre-processing are presented in Table~\ref{table:datasets}. 

\subsection{MovieLens-1M}

MovieLens dataset contains about 1 million explicit ratings users selected for movies~\cite{harper2016movielens}. We converted explicit ratings to implicit ones simply by assuming that a user watched a movie if she put a rating for it, otherwise she has not watched. We also ignore timestamps in all subsequent experiments. User features are the following: \textit{Age}, \textit{Gender}, \textit{Occupation} and \textit{ZipCode}. We converted the ZipCodes into US regions (e.g. \textit{MidWest}, \textit{South} etc.), therefore all user features are categorical with small cardinality. Item features are much richer. Each item is described by 1128 real numbers from interval [0,1]. Each value correspond to the strength of some tag. Examples of tags are (\textit{atmospheric}, \textit{thought-provoking}, \textit{realistic}, etc.). After excluding ratings which does not have both item and user features we have 914676 interactions in total. More statistics are provided in Table~\ref{table:datasets}. 

\subsection{BookCrossings}

BookCrossings dataset is a dataset scraped by~\cite{ziegler_2005} from the popular books rating web-site. Dataset contains both explicit and implicit ratings. At first, we have discarded all 0 ratings, then we have substituted all positive ratings with 1. Hence, we made implicit ratings from the explicit ones similarly to MovieLens pre-processing. We also selected 2999 most popular items and left only interactions with them. the amount of items is taken to be close to other datasets. It makes results more comparable and reduce computational workload. User feature include only \textit{Age} and \textit{Location}. We discard all the users with empty or too high age, then we have formed age groups of 10 years intervals. Location typically consist of town, region and country. We cleaned it as much as possible, but anyway it is a high cardinality categorical feature. There are 4 book features: \textit{Book-Title}, \textit{Book-Author}, \textit{Year-Of-Publication}, \textit{Publisher}. We extracted the key-words from titles and use them as features. All book features except the Year-Of-Publication have high cardinality. Parameters of pre-processed dataset are given in Table~\ref{table:datasets}.

\subsection{In-house Production Dataset}

The in-house production dataset consists of a data snapshot extracted from the database that contains user view events (interactions).
It is the largest dataset we experimented with. We did not filter out users or items if the amount of events with those is small. Hence, many users have very few interactions which makes this datasets challenging for Collaborative filtering methods like FCF. User features have several categorical features and some of those have high cardinality. In general, user features are similar to the user features of public datasets. Item features are similar to the tags features of the MovieLens data although not so reach. Further statistics are in Table~\ref{table:datasets}.

Particularly, FED-MVMF integrates user and item features for matrix factorization. We treat all user and item features as categorical (except for Movielens item-features, which are described by real numbers). Some features may have high cardinality (e.g. key-words of a book title) and different number of features per item. For instance, one book has 1 key-word in its title while another may have 3. Therefore, we processed all user (or item) features using a hash function of a certain output dimension. We call it hash size. This size depends on a dataset and exact numbers are provided in Table~\ref{table:datasets}. More precisely, we form stings like this \textit{\{feature\_name\}}\_\_{\{feature\_value\}} and hash all the strings of a user (item) into a vector of hash size. Originally, this is a vector of zeros. Hashing here means setting to 1 the corresponding coordinate of the vector. If there is a hashing collision we increase the corresponding value by 1. As a result, we obtain a sparse vector of fixed size for each user and item.
\begin{table}[h]
	\small
	\centering
	\resizebox{\columnwidth}{!}{%
		\centering
		\begin{tabular}{p{4cm}|ccccp{2cm}p{2cm}}
			\toprule
			Dataset & \# Users & \# Items & \# Interactions & Sparsity (\%) & User features & Item features\\
			& &  & $\R$ & & $\X$ & $\Y$\\
			\midrule
			Movielens & 6040 & 3064  & 914676 & 4.9\% & 3434 & 1128\\
			BookCrossings & 19912 & 2999 & 72794 & 0.12\% & 7405 & 10000\\
			In-house production dataset (anonymized) & 815614 & 3912 & 2213122 & 0.07\% & 300 & 300\\
			\bottomrule
	\end{tabular}}
	\caption{Overview of the datasets used in the study, where \# interactions refers to the total number of user-item interactions and Sparsity (\%) denotes the percentage of observed interactions in a particular dataset. User features represents the number of features used to encode personal data, while Item features is the number of features used to describe item features. In this study, the user-interactions, user features and item features are denoted by $\R,\ \X$ and $\Y$ respectively.}
	\label{table:datasets}
\end{table}

%% file: experiments_results.tex
\section{EXPERIMENTS AND RESULTS}\label{sec:exp_results}
We used Federated Collaborative Filter (FCF)~\cite{ammad2019federated,chai2019secure,dolui2019poster} as a baseline method. FCF is a federated matrix factorization method, however, does not integrate side-information sources. 

\paragraph{\textbf{Federated learning production system design:}}
We implement a production equivalent client-server architecture ~\cite{SHARMA201516},~\cite{Gamma95}, in which numerous clients are served by a single FL server and an Items service. All entities are implemented with Python (3.7.3) in a multiprocessing setup and cloud hosted on Ubuntu Linux 18.04 server infrastructure. The FL server has two data persistence layers of Redis (4.0.9) and PostgreSQL (9.6.10) databases, the Items service has only the Redis layer and the clients have no persistence layer. The hardware specifications are 8 cores with 64 gigabytes (GB) of memory for FL server and 16 cores with 16 GB of memory for Items service and clients.     

Both FL server and Items service use Gunicorn (19.7.1), the Python based web server to expose Application Programming Interfaces (API) as the only mechanisms to consume their services. Nginx (1.14) server lies between Gunicorn and clients optimizing service requests/responses by caching, compressing and decompressing payloads. Additionally, the FL server Base64 encodes/decodes all outgoing and incoming communications. 

The FL server initializes a master model for each of the available models with their respective hyper-parameters. FL server and clients query the Items service for item metadata. Thus, data on each client is formulated from the user’s personal data, item interactions and features associated with that item. Each client downloads a copy of an initialized master model and it's \textit{update-signature} from the FL server to their local storage. Model updates, metrics and inferences are derived from this local copy against user’s data on the client. Periodically, clients encapsulate an update payload by combining their model updates and performance metrics. The update payload is randomly uploaded to the FL server for aggregation. This random upload strategy differs from ~\cite{bonawitz2019towards} approach that maintains a subset of pre-selected known clients to upload their updates, by providing client anonymity, enhancing their privacy.   

The FL server implements queuing strategy to process incoming clients’ requests, responses and updates, for models at different stages. An update processor validates client payloads based on their update-signature before appending them to their  respective queues. A First In First Out (FIFO) model aggregator pops the oldest payload from the queue, recovering its model update and metrics. Updates are aggregated, an update counter incremented and the metrics persisted to a Structured Query Language (SQL) database for performance monitoring. The update counter is governed by the threshold parameter $\Theta$, defined for each model at initialization. When sufficient $\Theta$ aggregates have been accumulated, the aggregator first invalidates the current update-signature, implicitly informing clients not to upload more payloads and prepare for an updated master model and update-signature. Then, a new model composed of the previous model and the its updates aggregate, is promoted replacing its predecessor master model. A new update-signature is also generated for the renewed master model, prior to flushing the payload queues and updating the update processor and aggregator validation values.

\paragraph{\textbf{Hyper-parameters, training and evaluation criteria:}}
\begin{table}[t]
	\small
	\centering
	\resizebox{1\columnwidth}{!}{%
		\begin{tabular}{@{}ccccccccccc@{}}
			\toprule
			& $K$ & $\Theta$ & $\alpha$ & $\lambda_{1}$ & $\lambda_{2}$ & $\beta_{1}$ & $\beta_{2}$ & $\epsilon$ & $\gamma$ \\ \midrule
			BO Bounds &	[2, 25] &	[100, 50000] &	[4, 110] &	[0.001, 0.099] & [0.1, 1] &	[0.1, 0.55] & [0.55, 0.99] & [1e-8, 0.05] & [0.1, 0.99] \\ \bottomrule
			\multicolumn{10}{c}{In-house Anonymized Dataset} \\
			FCF & 25 & 32000 &	110 & -- & 1 &	0.1 & 0.99 & 1e-8 & 0.1 \\
			FED-MVMF & 22 & 47200 & 110 & 0.099 & 0.1 & 0.55 & 0.55 & 0.05 & 0.99 \\ \bottomrule
			\multicolumn{10}{c}{Movielens Dataset} \\
			FCF & 25 &	100 & 4 & -- & 1 &	0.1 & 0.99 & 1e-8 & 0.1 \\
			FED-MVMF & 25 & 3700 &	4 &	0.0989 & 1 & 0.1 & 0.98 & 0.0499 & 0.1\\ \bottomrule
			\multicolumn{10}{c}{Book Crossings Dataset} \\
			FCF & 25 & 10000 &	110 & -- & 1 &	0.1 & 0.55 & 0.05 & 0.99 \\
			FED-MVMF & 23 & 13300 &	4 &	0.099 & 0.1 & 0.1 & 0.55 & 1e-8 & 0.1 \\ \bottomrule				
		\end{tabular}
	}
	\caption{Summary of the hyper-parameters selected using Bayesian Optimization. Particularly, $K$, $\alpha$, $\lambda_{1}$ and $\lambda_{2}$ are defined for the model, while $\beta_{1}$ $\beta_{2}$, $\epsilon$ and $\gamma$ for Adam's learning rate and $\Theta$ is FL hyper-parameter defining the threshold on the amount of federated model-updates needed to update the master model.}
	\label{table:hyper-parameters}	
\end{table}

The FCF and FED-MVMF models share similar set of hyper-parameters except $\lambda_{1}$ which controls the strength of information shared with the side-information sources, and is specific to FED-MVMF. To choose optimal hyper-parameters, we used Bayesian optimization approach~\cite{snoek2012practical}. Table~\ref{table:hyper-parameters} illustrates the configuration settings for Bayesian optimization describing hyper-parameter bounds that were used to obtain the optimal set of hyper-parameters for the models. We performed 3 rounds of model rebuilds in production settings. In each round, the item interactions for every user were randomly divided into 80\% training and 20\% test sets, and select the metric value when 1000 iterations of federated model updates are reached, to ensure model convergence. Notably, in each federated iteration only a subset of users contribute to update the master model and report their performance metrics. Hence, at $Iteration=1000$ we take average of the previous 10 values to account for sampling biases in metric values.   

To evaluate the models, we use the widely adapted recommendation metrics \cite{BOBADILLA2013109}, Precision, Recall, F1, Mean Average Precision (MAP) and Normalized Mean Ranking (NMR) for the top $k=10$ predicted recommendations. The metrics were further normalized by the theoretically best achievable metrics for each dataset, to make them comparable.
\begin{align}
precision@k &  =  \frac{t_p^k}{t_p^k + t_n^k} & \label{eqn:precision}\\
\\
recall@k &  =  \frac{t_p^k}{t_p^k + f_n^k} & \label{eqn:recall}\\
\\
F1@k &  =  \frac{2*precision*recall}{precision + recall} & \label{eqn:f1} \\
\\
MAP@k & = \frac{1}{\|U_{test}\|} \sum\limits_{u=1}^{\|U_{test}\|} \frac{1}{\|I_{test}^u\|} \sum\limits_{i=1}^{k} \frac{1}{k} t_p^k [t_p^k - t_p^{(k-1)}  ]
\\
NMR & = \frac{1}{\|U_{test}\|} \sum\limits_{u=1}^{\|U_{test}\|} \frac{1}{\|I\|} \sum\limits_{i=1}^{\|I_{test}^u\|} rank_i
\end{align}

notation $t_p^k$, $t_n^k$, $f_n^k$ means true positives, true negatives and false negatives respectively, at ranked prediction list from positions $[1..k]$ and $U_test$ is the amount of data points in the test-set of a user. Note in all these metrics the possible values is in the range $[0, 1]$ where values closer to $1$ imply a better performance value except \textbf{NMR} the closer the value is to $0$ the better.

When making a comparison of metric values between for example the FCF and FED-MVMF model the "$\textbf{Impr \%}$" is quoted. In general this is given by the following.
\begin{equation}
\textbf{Impr \%} \ = \ 100 \times \frac{\text{Metric Mean (FED-MVMF)} - \text{Metric Mean (FCF)}}{\text{Metric Mean (FCF)}}
\label{eqn:impr}
\end{equation} 
where Metric Mean (FED-MVMF), Metric Mean(FCF) refer to the average value of the metric over a given number of model builds. This definition implies that if the $\textbf{Impr \%}$ value is positive then the FED-MVMF metric value is higher than the FCF value and vice-versa.
Next we show performance plots for these metrics demonstrating stable model convergence over several rounds of federated iterations.

\begin{figure*}[h]
	\centering
	\begin{minipage}{.18\columnwidth}
		\centering
		\includegraphics[width=\columnwidth]{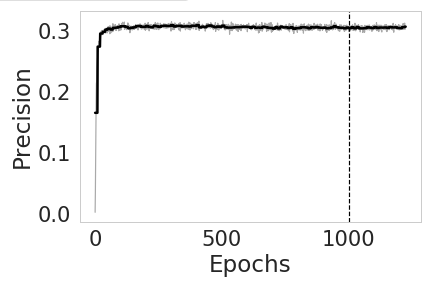}
	\end{minipage}
	\begin{minipage}{.18\columnwidth}
		\centering
		\includegraphics[width=\textwidth]{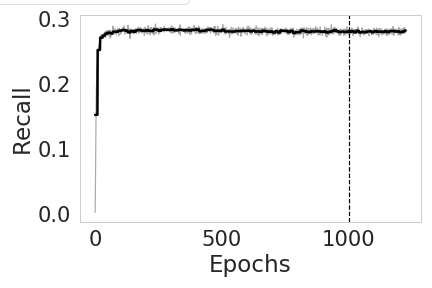}
	\end{minipage}
	\begin{minipage}{.18\columnwidth}
		\centering
		\includegraphics[width=\textwidth]{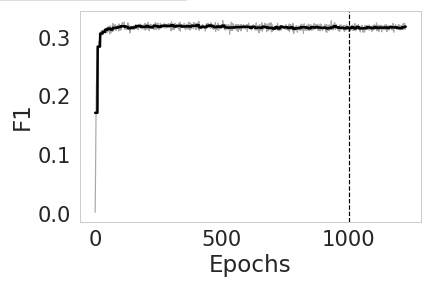}
	\end{minipage}
	\begin{minipage}{.18\columnwidth}
		\centering
		\includegraphics[width=\textwidth]{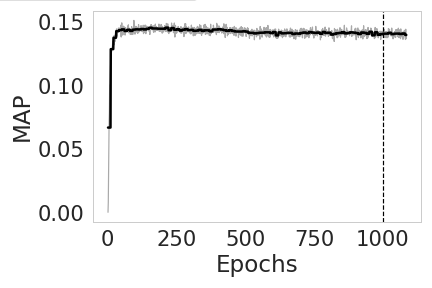}
	\end{minipage}
	\begin{minipage}{.18\columnwidth}
		\centering
		\includegraphics[width=\textwidth]{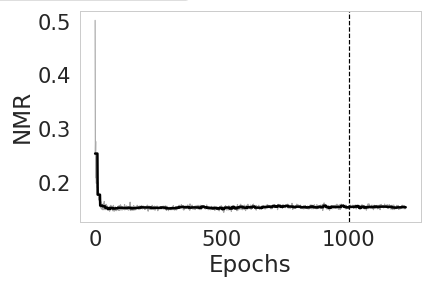}
	\end{minipage}

	\begin{minipage}{.18\columnwidth}
		\centering
		\includegraphics[width=\columnwidth]{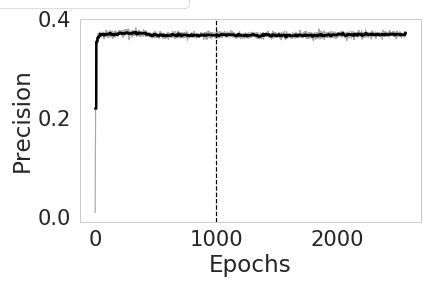}
	\end{minipage}
	\begin{minipage}{.18\columnwidth}
		\centering
		\includegraphics[width=\textwidth]{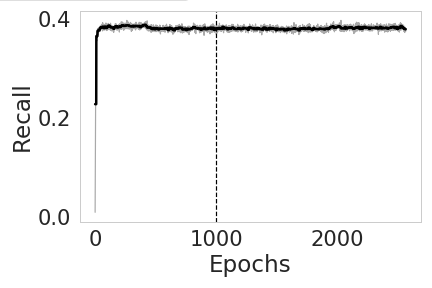}
	\end{minipage}
	\begin{minipage}{.18\columnwidth}
		\centering
		\includegraphics[width=\textwidth]{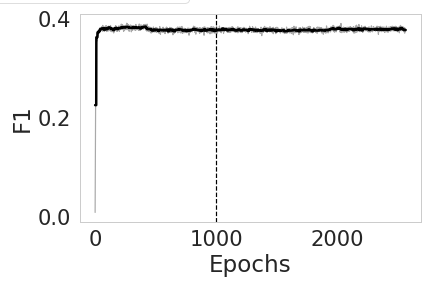}
	\end{minipage}
	\begin{minipage}{.18\columnwidth}
		\centering
		\includegraphics[width=\textwidth]{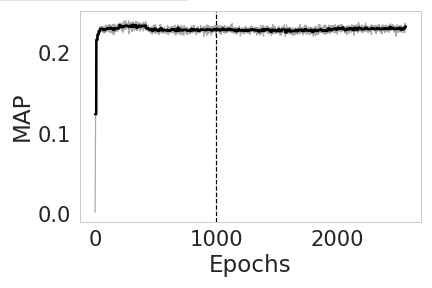}
	\end{minipage}
	\begin{minipage}{.18\columnwidth}
		\centering
		\includegraphics[width=\textwidth]{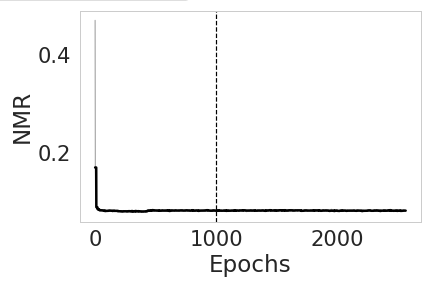}
	\end{minipage}

	\begin{minipage}{.18\columnwidth}
		\centering
		\includegraphics[width=\columnwidth]{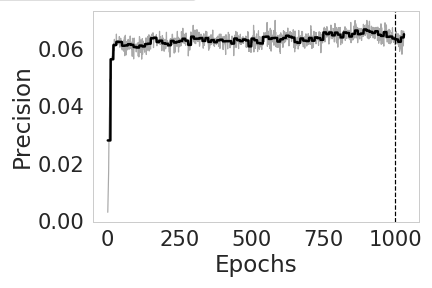}
	\end{minipage}
	\begin{minipage}{.18\columnwidth}
		\centering
		\includegraphics[width=\textwidth]{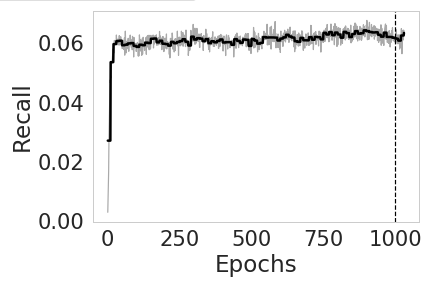}
	\end{minipage}
	\begin{minipage}{.18\columnwidth}
		\centering
		\includegraphics[width=\textwidth]{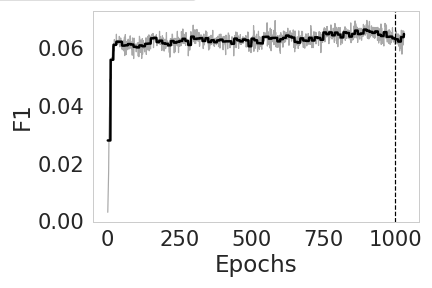}
	\end{minipage}
	\begin{minipage}{.18\columnwidth}
		\centering
		\includegraphics[width=\textwidth]{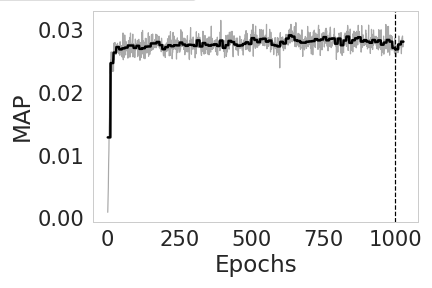}
	\end{minipage}
	\begin{minipage}{.18\columnwidth}
		\centering
		\includegraphics[width=\textwidth]{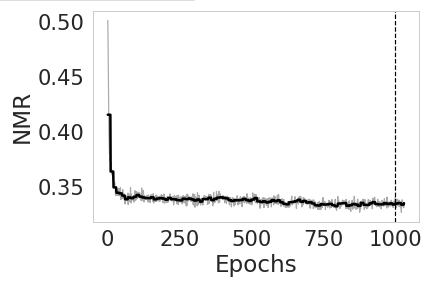}
	\end{minipage}
	\caption{Recommendation performance of FED-MVMF over several rounds of master model updates ($Iterations$). The in-house, Movielens and BooksCrossing results are shown in top-row, middle-row and bottom-rows respectively. The results demonstrate stable model convergence over several rounds of updates. In each federated iteration only a subset of users contribute to update the master model and report their performance metrics. Hence, at $Iteration=1000$ we take average of the previous 10 values to account for sampling biases in metric values.}
	\label{fig:perf_plots_fed_mvmf}
\end{figure*}

\paragraph{\textbf{Convergence Analysis and Performance Metric Selection:}}
Figure~\ref{fig:perf_plots_fed_mvmf} shows the FED-MVMF model's convergence for each of the data sets. In the figure, Y-axis denotes the recommendation metric, whereas X-axis represents $Iterations$ or rounds of master model updates. At $Iteration=1000$ we compute average of previous 10 values and report in Table~\ref{table:ws_performances}.

\paragraph{\textbf{Recommendation Performance:}}
We first compare the performance of the proposed FED-MVMF with FCF on three real personalized recommendation data sets. The results demonstrate that the FED-MVMF method outperforms FCF substantially in most cases as shown in Table~\ref{table:ws_performances}. Importantly, the relative improvement in performances goes upto 70\% when measured across different metric and datasets. We noticed a significant improvement for highly sparse datasets such as the in-house anonymized production dataset and BooksCrossing dataset. 

This finding implies that the use of side-information sources is beneficial for production datasets which are inherently sparse in nature. The strength of predictive signal could be dependent on the type and nature of side-information sources used for a particular dataset. The results demonstrate larger improvements in recommendation performances of in-house and BookCrossing dataset that integrated discretized features compared to the Movielens dataset which included dense features encoded with real-values. Here, our main research goal was to propose a FED-MVMF method that can take advantage of the side-information sources in a federated learning. The development of a multi-view model is a technically challenging research problem owing to the federated nature of the multi-view data sets and hard constraints of the federated learning design. Our results confirmed that the developed solution takes advantage of the side-information sources to provide substantially improved recommendations.      
\begin{table}[t]
	\small
	\centering
	\resizebox{1\columnwidth}{!}{%
		\begin{tabular}{@{}cccccc@{}}
			\toprule
			& Precision & Recall & F1 &	MAP & NMR\\ \midrule
			\multicolumn{5}{c}{In-house Anonymized Dataset} \\
			FCF & 	0.1811$\pm$0.0009 &	0.1816$\pm$0.0007 &	0.1812$\pm$0.0007 &	0.0842$\pm$0.0009 &	0.3097$\pm$0.0017\\
			FED-MVMF &	0.2771$\pm$0.0022 &	0.2779$\pm$0.0028 &	0.277$\pm$0.0023 & 0.1411$\pm$0.0021 &	0.1545$\pm$0.0006\\
			Impr (\%) &	53 & 53 & 53 & 68 & 50\\ \bottomrule
			\multicolumn{5}{c}{Movielens Dataset} \\
			FCF & 0.3410$\pm$0.0100 & 0.3571$\pm$0.0019 & 0.3525$\pm$0.0037 & 0.2055$\pm$0.0090 & 0.1006$\pm$0.0021 \\
			FED-MVMF & 0.3666$\pm$0.0017 & 0.3825$\pm$0.0043 & 0.3785$\pm$0.0037 & 0.2295$\pm$0.0013 & 0.0841$\pm$0.0012 \\
			Impr (\%) & 8 & 7 & 7 & 12 & 16 \\ \bottomrule
			\multicolumn{5}{c}{Book Crossings Dataset} \\
			FCF & 0.0431±0.0025 & 0.0428±0.0023 & 0.0431±0.0024 & 0.0166±0.0012 & 0.39±0.0015 \\
			FED-MVMF & 0.0639±0.0011 & 0.0625±0.0016 & 0.0636±0.0013 & 0.0284±0.0012 & 0.3378±0.006 \\
			Impr (\%) & 48 & 46 & 48 & 71 & 13 \\ \bottomrule				
		\end{tabular}
	}
	\caption{Comparison of the test set performance between FED-MVMF and FCF methods. The values denote the mean $\pm$ standard deviation of metric values across 3 different model builds. Impr (\%) refers to the relative percentage improvement between the mean values of FED-MVMF and FCF. FED-MVMF model outperforms the FCF model showing a substantial improvement, going upto 70\%.}
	\label{table:ws_performances}	
\end{table}
Furthermore, we also present payload estimates in terms of model sizes and time, particularly when tested in production settings. 

\paragraph{\textbf{Payload Analysis:}}
We next analyzed the payloads for the two federated models. The item-factor matrix $\Q$ is common between FCF and FED-MVMF models, however
an additional payload for the FED-MVMF model comes from the user-features factor matrix $\Ub$. The size of $\Q$ depends on number of latent factors $K$ and number of items $N_v$, whereas the size of $\Ub$ is dependent on $K$ as well as the number of user-features $D_u$. In the case of FED-MVMF, both $\Q$ and $\Ub$ are transmitted as part of model updates between the FL clients and the FL server. As expected, we observed increased payloads for the FED-MVMF model compared to the FCF which does not include user-features. Notably, the model sizes scales linearly with increasing number of items and user-features. The relative increase in model size ranges from 80\% to 200\% across the three data sets, whereas computation time increases from 24\% to 52\%. We think that for the case of Books-Crossing, the larger time taken by FCF to compute model updates could be merely a technical artifact and need further clarification. The model update time on FL server also includes the time taken by Item-Server to update item-feature matrix $\V$ and compute gradients for $\Q$.  

Compared to the FCF model, the proposed FED-MVMF model increases payloads for the FL recommendation system. However FED-MVMF yields better recommendation performances and additionally provides principled solution to cold-start recommendation problems in FL.
\begin{table}[t]
	\small
	\centering
	\resizebox{1\columnwidth}{!}{%
		\begin{tabular}{@{}ccccccc@{}}
			\toprule
			& \multicolumn{5}{c}{\textbf{FL Client}} & \textbf{FL Server} \\ \midrule
			& \multicolumn{2}{c}{Model Download}          & Model Update         & \multicolumn{2}{c}{Model Upload}            & Model Update         \\ \midrule
			& Size (KB)            & Time (MS)            & Time (MS)            & Size (KB)            & Time (MS)            & Time (MS)            \\ \midrule
			\multicolumn{7}{c}{In-house Anonymized Dataset}                 \\ 
			FCF & 265.28 &	7.39 &	9.47  &	242.31 & 7.35 &	2.87
			\\
			FED-MVMF &  483.16 & 10.78 & 11.78 & 436.07  &	9.06 &	4.19
			\\
			Impr (\%) &  82 &	45 & 24 & 79  &  23 &	45
			\\ \midrule
			\multicolumn{7}{c}{Movielens Dataset}                 \\ 
			FCF &  402.17 &	8.06 &	10.46 &	402.17 &	16.09 &	3.44
			\\
			FED-MVMF & 849.47 &	14.80 &	15.98 &	849.38 &	15.29 &	3.30
			\\ 
			Impr (\%) &  111 &	83 & 52  & 111 &	-4 &	-4
			\\ \midrule
			\multicolumn{7}{c}{Book Crossings Dataset}                 \\ 
			FCF &  390.84 &	7.23 &	46.56 &	350.85 &	4.62 &	2.95
			\\
			FED-MVMF & 1021.80 &	15.78 &	17.40 &	1057.00 &	7.07 &	3.16
			\\  
			Impr (\%) &  161 &	118 & -62 &	201 &	53 &	7
			\\ \bottomrule 
		\end{tabular}%
	}
	\caption{Payload comparison between FCF and FED-MVMF in terms of the model size (KB=KiloBytes) and time (MS=Milliseconds). The FL client downloads the model (Model Download), update the local model and computes the master model updates or gradients (Model Update) and uploads the gradients to the FL Server (Model Upload). The FL server aggregates the updates arriving from the clients and updates the master model (Model Update).}
	\label{tab:payloads}
\end{table}

\paragraph{\textbf{Cold-start recommendation performance:}}
FED-MVMF provides a principled solution to the commonly occurring problem in production: cold-start recommendations. To demonstrate the usefulness of the FED-MVMF model for cold-start predictions, we conducted comprehensive analysis of all the three cold-start scenarios: cold-start users, cold-start items and cold-start user-items. 
For case of cold-start users scenario ,the model did not observe any of the interaction data. A random subset of 10\% users were completely held-out during the model training and model parameters were learned with remaining 90\% of the users.  For case of cold-start items, a random subset of 10\% items were entirely left-out during the model training and model parameters were learned with remaining 90\% of the items. For case of cold-start users-items, a random subset of 10\% users and items were excluded from the model training and model parameters were learned with remaining 90\% of the users and items. Likewise, 3 rounds of the model rebuilds were done for each of the scenario. Table~\ref{table:cs_performances} illustrates the recommendation performances across all scenarios. 

The result demonstrate that without loss of generality, the FED-MVMF model can be used for cold-start recommendations reliably. Specifically, the model shows good cold-start prediction performances for a new user, which is a fundamentally valuable in a federated learning solution where new users are enrolled in the service continuously. The performance of cold start item prediction is observed to be lower than that of cold start user indicating that prediction may be improved further. It is likely that the difference is due to lower quality of the item side-information source. Moreover, the low standard deviation of the results indicates that model predictions are precise across variations in training sets. 
\begin{table}[t]
	\small
	\centering
	\resizebox{1\columnwidth}{!}{%
		\begin{tabular}{@{}cccccc@{}}
			\toprule
			& Precision & Recall & F1 &	MAP & NMR\\ \midrule
			\multicolumn{5}{c}{In-house Anonymized Dataset} \\
			CS-Users &	0.3559$\pm$0.0015 &	0.3359$\pm$0.0012 & 0.3518$\pm$0.0014 & 0.1743$\pm$0.0012 & 0.0621$\pm$0.0025\\
			CS-Items &	0.0263$\pm$0.0006 &	0.0515$\pm$0.0011 &	0.0292$\pm$0.0007 & 0.0114$\pm$0.0003 &	0.2727$\pm$0.0032\\
			CS-Users-Items &	0.1739$\pm$0.0012 &	0.3352$\pm$0.0027 &	0.1916$\pm$0.0014 & 0.1384$\pm$0.0030 &	0.0792$\pm$0.0030\\ \midrule
			\multicolumn{5}{c}{Movielens Dataset} \\
			CS-Users &	0.4618$\pm$0.0086 &	0.5008$\pm$0.0102 &	0.4984$\pm$0.0093 & 0.3504$\pm$0.0088 &	0.3025$\pm$0.0072 \\
			CS-Items &	0.0043$\pm$0.0003 &	0.0464$\pm$0.0032 &	0.0291$\pm$0.002 & 0.0031$\pm$0.0005 &	0.4157$\pm$0.0006\\
			CS-Users-Items & 0.0440$\pm$0.0031 & 0.4528$\pm$0.0252 & 0.2903$\pm$0.0173 & 0.0239$\pm$0.0015 & 0.3384$\pm$0.0084\\ \midrule
			\multicolumn{5}{c}{Book Crossings Dataset} \\
			CS-Users &	0.0521$\pm$0.0027 &	0.0559$\pm$0.0022 &	0.0531$\pm$0.0024 & 0.0254$\pm$0.0032 &	0.3396$\pm$0.0034 \\
			CS-Items &	0.0054$\pm$0.0005 &	0.012$\pm$0.001 &	0.0063$\pm$0.0005 & 0.0047$\pm$0.0004 &	0.4918$\pm$0.0128\\
			CS-Users-Items &	0.0166$\pm$0.0030 &	0.0399$\pm$0.0068 &	0.0199$\pm$0.0034 & 0.0137$\pm$0.0054 &	0.3503$\pm$0.0055\\  \bottomrule				
		\end{tabular}
	}
	\caption{Cold-Start recommendation performance metrics of FED-MVMF using different metrics averaged over all users. The values denote the mean $\pm$ standard deviation across 3 different model builds. The proposed FED-MVMF model made possible to recommend items for challenging cold-start scenarios in federated learning model.}
	\label{table:cs_performances}	
\end{table}

%% file: conclusion.tex
\section{CONCLUSION}
We introduced the federated multi-view matrix factorization method where the federated paradigm does not require collecting raw user data to a centralized server thus enhancing the user privacy. The proposed federated multi-view model is tested on three different datasets and we showed that including the side-information from both users and items increases recommendation performance compared to a standard federated Collaborative Filter. The multi-view approach provides a solution to the cold-start problem common to standard Collaborative Filter recommenders. The results establish that the federated multi-view model can provide better quality of recommendations without comprising the user's privacy in the widely used recommender applications.
\paragraph{\textbf{Future Work:}} An important aspect of any federated learning system is the amount of data, or the payload, to be moved between the FL server and user. In any matrix factorization based recommender system the model size, or factor matrix size is directly proportional to the number of items to recommend which in large federated recommendation systems is not feasible. Our main challenge is to break the direct dependence of model size on the number of items to recommend.